\RequirePackage{fix-cm}
\documentclass[smallextended]{svjour3}       
\smartqed  

\usepackage[english]{babel}
\usepackage{soul}
\usepackage{lmodern} 

\usepackage{comment}
\usepackage{url}
\usepackage{hyperref}

\usepackage[authoryear]{natbib}

\usepackage[most]{tcolorbox}

\usepackage{soul}
\usepackage{tabularx}
\usepackage{amsmath,amssymb,amsfonts}
\usepackage{algorithmic}
\usepackage{graphicx}
\usepackage{textcomp}
\usepackage{xcolor}
\usepackage{comment}
\usepackage{multirow} 
\usepackage{footnote}
\usepackage{url}
\usepackage{multirow}
\usepackage{makecell}
\usepackage{multirow}
\usepackage{subcaption}

\makeatletter
\renewcommand{\figurename}{\textbf{Fig.}}

\long\def\@makecaption#1#2{%
  \vskip\abovecaptionskip
  \sbox\@tempboxa{\textbf{#1}~#2}%
  \ifdim \wd\@tempboxa >\hsize
    \textbf{#1}~#2\par
  \else
    \global \@minipagefalse
    \hb@xt@\hsize{\hfil\box\@tempboxa\hfil}%
  \fi
  \vskip\belowcaptionskip}
\makeatother

\usepackage{amsmath} 
\usepackage{caption}

\usepackage[title]{}
\usepackage{tabularx}
\usepackage{booktabs}
\usepackage{longtable}
\usepackage{hyperref}

\usepackage{svg}
\usepackage{enumitem}
\usepackage{listings}
\usepackage{color}
\usepackage{float}

\definecolor{verylightgray}{gray}{0.97}

\usepackage{pgfplots}
\usepgfplotslibrary{groupplots}
\usetikzlibrary{patterns,shapes.arrows}

\pgfplotsset{compat=1.18}

\usepackage[utf8]{inputenc}

\begin{document}

\title{Advancing Dialectal Arabic to Modern Standard Arabic Machine Translation
}
%



\author{Abdullah Alabdullah    \and Lifeng Han \footnote{Corresponding Author} \and Chenghua Lin
}


\institute{AA \at
              Primary Affiliation: University of Manchester, Manchester M13 9PL, The United Kingdom \\
              \email{abdullah.alabdullah@student.manchester.ac.uk }           \\
            \emph{Current address:} School of Informatics, University of Edinburgh, The United Kingdom \\
            Active \email{a.alabdullah@sms.ed.ac.uk} 
           \and
           LH \at 
                University of Manchester, Manchester M13 9PL, The United Kingdom \\
              LIACS and LUMC, Leiden University, Leiden, NL \\ \email{l.han@lumc.nl}
            \and 
            CL \at 
            University of Manchester, Manchester M13 9PL, The United Kingdom \\
              \email{chenghua.lin@manchester.ac.uk} 
}

\date{Received: date / Accepted: date}

\maketitle

\begin{abstract}

Dialectal Arabic (DA) poses a persistent challenge for natural language processing (NLP), as most everyday communication in the Arab world occurs in dialects that diverge significantly from Modern Standard Arabic (MSA). This linguistic divide impedes progress in Arabic machine translation.
This paper presents two core contributions to advancing DA–MSA translation for the Levantine, Egyptian, and Gulf dialects, particularly in \textit{low-resource} and \textit{computationally constrained} settings: (i) a comprehensive evaluation of training-free prompting techniques, and (ii) the development of a resource-efficient fine-tuning pipeline.
Our evaluation of \textbf{prompting} strategies across \textit{six} large language models (LLMs) found that few-shot prompting consistently outperformed zero-shot, chain-of-thought, and our proposed Ara-TEaR method. Ara-TEaR is designed as a three-stage self-refinement prompting process, targeting frequent meaning-transfer and adaptation errors in DA–MSA translation. In this evaluation, GPT-4o achieved the highest performance across all prompting settings.
For \textbf{fine-tuning} LLMs, a quantized Gemma2-9B model achieved a chrF++ score of 49.88, outperforming zero-shot GPT-4o (44.58). Joint multi-dialect trained models outperformed single-dialect counterparts by over 10\% chrF++, and 4-bit quantization reduced memory usage by 60\% with less than 1\% performance loss.
The results and insights of our experiments offer a practical blueprint for improving dialectal inclusion in Arabic NLP, showing that high-quality DA–MSA machine translation is achievable even with limited resources and paving the way for more inclusive language technologies.


\end{abstract}

\section{Introduction}

Arabic is the fifth most spoken language globally, with over 334 million speakers \citep{eberhard2023ethnologue}. This vast linguistic landscape is marked by a significant linguistic divide: Modern Standard Arabic (MSA) serves as the formal register, while numerous dialectal varieties, such as Egyptian, Levantine, and Gulf, dominate everyday communication. These dialects diverge considerably from MSA in vocabulary, style, and syntax, posing substantial barriers to language technologies. Existing natural language processing (NLP) systems, while effective for MSA, often fail when faced with dialectal Arabic (DA) content \citep{mohamed2023alphamwe_Arabic}. For example, social media monitoring tools, built primarily for MSA, struggle to interpret sentiment and meaning in dialectal content, which is widely used in everyday communication and increasingly prevalent in social media text \citep{alhafni2024codafication}.

Despite the rapid progress of Large Language Models (LLMs), research specifically applying them to effectively translate dialectal Arabic into Modern Standard Arabic (DA--MSA) remains notably limited. Benchmarking efforts like AraDiCE \citep{mousi2024aradice} highlight persistent gaps in dialect translation and comprehension.
Although some efforts have focused on DA-MSA MT methods, there remains a critical need for two specific areas of investigation: (1) a comprehensive and systematic evaluation of diverse \textit{prompting techniques} across various LLMs for DA-MSA translation, and (2) a detailed study featuring a systematic experimentation process to identify performance-critical variables affecting translation quality when \textit{fine-tuning} LLMs for low-resource DA-MSA MT. 


This paper addresses these gaps by extensively exploring and evaluating two distinct approaches for DA-MSA neural machine translation (NMT) within low-resource and computationally constrained environments. The first approach investigates training-free prompting strategies, systematically assessing how different prompt designs and contextual cues impact translation quality across various dialects and LLMs without the need for model fine-tuning. The second approach focuses on developing resource-efficient fine-tuning methodologies that achieve high-quality DA-MSA translation while significantly reducing computational requirements. This involves exploring techniques such as Low-Rank Adaptation (LoRA), quantization, data augmentation, and multi-dialect joint training, with systematic comparisons presented to identify the trade-offs between translation quality and efficiency.

To support future research and development, our data and code are openly shared at Github \footnote{\url{https://github.com/er-abd/Advancing-DA2MSA-MT}}. Our fine-tuned models can be downloaded from Huggingface \footnote{\url{https://huggingface.co/collections/er-abd/}} (advancing-da2msa-mt).

\section{Background and Related Work}\label{sec:background}

\subsection{Traditional Approaches for DA-MSA translation}
Before the rise of neural models, DA--MSA translation relied heavily on rule-based and statistical techniques. Early systems incorporated hand-crafted morphological analyzers, lexicons, and linguistic transformation rules tailored to specific dialects. Tools like the Buckwalter Morphological Analyzer \citep{buckwalter2004} and MADAMIRA \citep{pasha2014madamira} were widely used to generate morphological analyses and standardize dialectal tokens. Rule-based normalization techniques also attempted to map dialectal forms into MSA using curated substitution lists or token-level alignments \citep{habash2005arabic}.

In addition to rule-based transformation, lexical mapping and transliteration strategies were explored. Dictionary-based lookup and transliteration tools sought to align phonetically similar tokens, but were hindered by high orthographic variability and lack of standardized spelling in dialects \citep{habash2012coda}.

Phrase-based statistical machine translation (SMT), as implemented in systems like Moses \citep{koehn2007moses}, represented a major step forward. SMT systems used probabilistic phrase alignments and required bilingual corpora to learn translation patterns. However, the scarcity of high-quality DA--MSA parallel corpora has limited SMT performance \citep{klementiev2012toward}. Researchers attempted to compensate by generating synthetic parallel data or leveraging existing corpora such as MADAR \citep{bouamor2018madar}, but the lack of broad dialectal coverage remained a bottleneck.

Despite their utility, traditional approaches struggled with out-of-vocabulary dialectal words, lacked the capacity to generalize across dialects, and could not automatically learn deep contextual mappings. These limitations motivated the shift toward data-driven neural approaches, which proved more effective in scaling across dialects and modeling a variation-rich language.

\subsection{Neural Machine Translation for DA-MSA translation}
Neural Machine Translation (NMT) significantly advanced the capabilities of DA-MSA translation systems, addressing challenges that traditional MT systems failed to handle such as the morphological richness, orthographic ambiguity, and dialectal variation of Arabic dialects~\citep{baniata2021transformer}. NMT revolutionized the field by leveraging deep neural networks capable of modeling non-linear and context-dependent relationships between dialectal inputs and MSA outputs, enabling more fluent and accurate translation across dialects ~\citep{vaswani2017attention}

A neural network (NN) is a multi‑layer architecture of interconnected neurons trained to learn representations from data, forming the foundation of modern language models~\citep{lecun2015deep}. NMT typically uses encoder–decoder architectures, where an encoder processes source text and a decoder generates the target translation. Compared to traditional methods, NMT offers better generalization to unseen inputs, learns complex dialect-to-MSA mappings without hand-crafted rules, and supports joint multilingual training across dialects~\citep{johnson2017google, baniata2021transformer}.




\subsection{Prompting and Fine-Tuning LLMs for DA–MSA Translation}

The decoder-only architecture has proven highly effective for MT, as it autoregressively generates each output token conditioned on prior tokens. While encoder–decoder models initially dominated, decoder-only models gained popularity for their scalability and strong generative performance at scale~\citep{caillaut2024scaling}. Most large language models (LLMs), including GPT and LLaMA, adopt this decoder-only Transformer architecture, trained on vast multilingual corpora to capture complex linguistic patterns~\citep{naik2024decoder}. The emergence of LLMs has transformed MT for low-resource languages, including the understudied task of DA–MSA translation. These models learn to predict the next token in a sequence based on the context, allowing them to generate coherent and contextually appropriate outputs across diverse linguistic inputs. Their ability to capture long-range dependencies and semantic nuances enables them to perform a wide range of NLP tasks, including translation, summarization, and question answering, with minimal task-specific training. Unlike traditional MT systems, LLMs can generalize across domains and dialects, enabling utilization through both model \textit{prompting} and \textit{fine-tuning}. Model prompting involves task-specific instructions without modifying the model’s internal parameters, while fine-tuning adapts the model by updating its weights on a labeled dataset. Prompting excels in scenarios where high-quality resources are absent, whereas fine-tuning is preferred for robust, domain-adapted performance using relatively small datasets~\citep{chen2023promptingtaxonomy}.

LLM prompting leverages in-context learning (ICL), where models generate task outputs based solely on examples and instructions provided at inference time, without parameter updates. This training-free paradigm relies on prompt engineering to achieve high-quality translations by framing tasks clearly and contextually~\citep{liu2023pre}. Effective prompts typically define the translation task, specify the source dialect, and may include exemplar translations (few-shot prompting) or structured reasoning instructions (Chain-of-Thought prompting) to guide the model~\citep{wei2022chain}.

Prompting has become a central theme in LLM research due to its scalability and simplicity~\citep{liu2023pre}. For low-resource MT tasks like DA--MSA translation, prompting is promising due to its minimal data requirements and capacity to generalize across underrepresented dialects~\citep{chung2022scaling}. Recent work has applied prompting to DA--MSA translation, including TARJAMAT~\citep{abdul2023tarjamat}, which showed LLMs outperforming many commercial systems in Arabic dialect-to-English translation. In NADI 2024~\citep{abdulmageed2024nadi}, Arabic Train achieved top performance using Jais with a combined zero/few-shot setup~\citep{harrat2024arabictrain}, while the winner in the OSACT 2024 shared task ~\citep{osact2024} used ChatGPT with 3-shot prompting and self-correction to surpass their fine-tuned baseline~\citep{atwany2024osact}. Nevertheless, comparative and comprehensive evaluations of various prompting strategies across different LLMs, model sizes, and dialects remain limited. 
Additionally, key factors such as the impact of dialect identification, contextual cues, and model scaling on translation quality remains underexplored, all of which are \textit{areas that we aim to investigate in our work}.

While prompt engineering guides model outputs, \textbf{fine-tuning} directly adapts a pre-trained model by updating its parameters on a labeled and task-specific dataset, thereby improving its performance through specialization~\citep{howard2018universal}. This process typically involves initializing an LLM with pretrained weights and continuing training on a smaller supervised corpus, allowing the model to internalize task-specific linguistic patterns and vocabulary~\citep{raffel2020exploring}, e.g. in MT \citep{yin2024cantonmt}. Fine-tuning is particularly suitable in low-resource settings like DA--MSA translation, as it enables models to adapt to specific domains with limited labeled data. 
Compared to prompting, which relies on inference-time instructions, fine-tuning often yields more robust and consistent results when even modest amounts of annotated data are available~\citep{liu2023pre,han2024neural}.

In Arabic domain, an early work by~\cite{khalifa2021self} demonstrated the benefits of fine-tuning for dialectal adaptation in foundational NLP tasks. More recently, Sirius\_Translators fine-tuned the encoder--decoder model AraT5 across five Arabic dialects~\citep{alahmari2024sirius}, while CUFE adapted the decoder-only LLaMA-3 model with strong results~\citep{ibrahim2024cufe}. Lahjawi-D2MSA, a fine-tuned version of the Kuwain-1.5B small language model, was proposed for translating 15 Arabic dialects to Modern Standard Arabic and achieved a high human-rated fluency score of 78\% in fluency evaluations~\citep{hossain2025lahjawi}. Despite these advancements, there is still a lack of systematic studies investigating the \textit{performance-critical variables that impact fine-tuning for DA--MSA MT}.

To address this, our work conducts a series of controlled ablation experiments isolating the effects of architecture, training strategy, and hyper-parameters. Additionally, we evaluate the effectiveness of paraphrasing-based data augmentation, joint-dialect training, and active learning. We further analyze the trade-offs of 4-bit quantization on translation quality and efficiency. Finally, we compare fine-tuned and zero-shot prompted models across multiple architectures and sizes, offering a comprehensive view of the performance and scalability trade-offs in DA--MSA LLM-based translation.

\section{Methodology} 
Our work investigates two approaches for enhancing the translation of DA--MSA under low-resource and computationally constrained environments:

\begin{itemize}
    \item \textit{Training-free prompting}, which leverages LLMs through strategically engineered prompts without altering model parameters.
    \item \textit{Resource-efficient fine-tuning}, which adapts mid-sized models using parameter-efficient methods, such as Low-Rank Adaptation (LoRA).
\end{itemize}

The rationale for employing a dual approach stems from both practical constraints and linguistic complexities, as discussed in Section ~\ref{sec:background}. Prompt-based methods yield competitive performance with minimal resource demands, making them suitable for settings where computational capabilities or annotated data are limited. However, such methods often face limitations in addressing the morphological complexity and dialectal variability inherent in Arabic, particularly when faced with highly informal, idiomatic, or structurally complex input not encountered during pre-training ~\citep{abdul2023tarjamat}. In contrast, fine-tuning enables models to learn directly from DA--MSA parallel data, thereby producing more robust and dialect-aware translations. However, this advantage comes with an increase in training overhead and computational costs. Nevertheless, it is important to note that the inference costs associated with larger, general-purpose models used in prompting approaches can lead to significantly higher carbon emissions per user request than smaller, fine-tuned models, even with similar parameter counts~\citep{luccioni2023power}. Given the methodological divergence of the two approaches, each is addressed through a tailored experimental setup, as outlined in ~\ref{sec:training-free-prompting} and ~\ref{sec:model_fine_tune_setup}

\subsection{Dataset Curation}
This study used a curated set of high-quality DA-to-MSA parallel corpora to support both prompting and fine-tuning experiments. The selected datasets span both formal and informal domains, introducing variation in linguistic style and syntactic complexity. This diversity aimed to enhance the robustness and generalizability of the model across a broad spectrum of translation scenarios. Our curated corpora encompassed three major Arabic dialects: Levantine (LEV), Gulf (GLF), and Egyptian (EGY).

The primary resource utilized was the Multi-Arabic Dialect Applications and Resources (MADAR) corpus \citep{bouamor2018madar}, comprising approximately 43,000 parallel sentences in its training split. This \textit{travel}-domain dataset, translated by professional translators, is widely recognized for its linguistic precision and consistency. MADAR has been extensively used in shared tasks such as OSACT ~\citep{osact2024} and NADI \citep{abdulmageed2024nadi}. To complement the formal domain focus of MADAR, the Dial2MSA-Verified corpus \citep{khered2025dial2msa} (19,775 training pairs) was also incorporated. This dataset features \textit{social media content} characterized by \textit{colloquial} and \textit{idiomatic} expressions, and was selected due to its native-speaker validation.

To further improve the models’ understanding of MSA, 5,000 MSA--MSA sentence pairs from the UFAL Parallel Corpus of North Levantine Arabic \citep{krubinski2023ufal} were included. These same-source-target examples served as a form of regularization, enabling the model to learn when no transformation is necessary, particularly relevant given the structural affinities between Arabic dialects and MSA. In addition, a synthetic dataset of 43,000 samples was constructed by prompting GPT-4o to paraphrase the MSA references from the MADAR training data. This augmentation aims to increase lexical and stylistic variation on the MSA side while preserving the original dialectal input. Paraphrasing was preferred over back-translation due to its capacity to enrich lexical diversity without altering the dialectal text, an important consideration given the morphological complexity of Arabic. Furthermore, manual testing revealed that existing language models underperformed in translating from MSA to Dialectal Arabic, making back-translation suboptimal.

Three training sets were developed, as detailed in Table~\ref{tab:dataset_composition}:

\begin{itemize}
    \item \textbf{Minimal-Gold} included only the MADAR training split.
    \item \textbf{Extended-Gold} extended the \textit{Minimal-Gold} dataset with Dial2MSA-Verified training data.
    \item \textbf{Extended-Gold-Silver} extended the \textit{Extended-Gold} (62,775 examples) with 43,000 GPT-4o–paraphrased MSA references and 5,000 UFAL MSA–MSA pairs, yielding a total of 110,775 training examples.
\end{itemize}

These sets ensured balanced dialectal representation while progressively increasing the scale and diversity of training data.

\begin{table}[htbp]
\centering
\begin{tabular}{lccccc}
\hline
Training Set & LEV & GLF & EGY & Total \\
\hline
\textbf{Minimal-Gold} & 17k & 13.8k & 12.2k & 43k \\
\textbf{Extended-Gold} & 21.1k & 20.4k & 21.3k & 62.8k \\
\textbf{Extended-Gold-Silver} & 38.1k & 34.2k & 33.5k & 110.8k \\
\hline\hline
\end{tabular}
\caption[Training Dataset Compositions Across Dialects]
{Breakdown of training datasets \textit{Minimal-Gold} , \textit{Extended-Gold} , and \textit{Extended-Gold-Silver}  by dialect and size}
\label{tab:dataset_composition}
\end{table}

For evaluation purposes, two test sets were constructed, as outlined in Table~\ref{tab:test_sets}:

\begin{itemize}
    \item \textbf{Small-Test} comprised 600 examples (200 per dialect), randomly sampled from the test split of MADAR.
    \item \textbf{Large-Test} comprised of 1,200 examples (400 per dialect), randomly sampled from the test split of both MADAR (70\%) and Dial2MSA-Verified (30\%) to ensure domain and linguistic variability.
\end{itemize}

\begin{table}[htbp]
\centering
\begin{tabular}{lcccc}
\hline
Test Set & LEV & GLF & EGY & Total \\
\hline
\textbf{Small-Test} & 200 & 200 & 200 & 600 \\
\textbf{Large-Test} & 400 & 400 & 400 & 1200 \\
\hline\hline
\end{tabular}
\caption[Evaluation Dataset Compositions]
{Breakdown of test datasets \textit{Small-Test} \& \textit{Large-Test} by dialect and size}
\label{tab:test_sets}
\end{table}

The two test sets might contain potential overlapping examples since both included randomly sampled examples from the test split of MADAR.

\subsection{Evaluation}

Translation quality in this study was assessed using two widely adopted automatic evaluation metrics: SacreBLEU and chrF++. SacreBLEU \citep{post-2018-call} provides a standardized and reproducible implementation of the original BLEU metric \citep{papineni-etal-2002-bleu}, which quantifies n-gram overlap between system-generated translations and references. Although BLEU-based metrics are prevalent in machine translation evaluation, their emphasis on surface-level lexical matches can limit their effectiveness for morphologically rich languages such as Arabic \citep{han2021translation}.

In contrast, chrF++ \citep{Popovic2017chrF++} computes F-scores over both \textit{character} and \textit{word}-level n-grams, enhancing its sensitivity to morphological variations and orthographic inconsistencies; features that are particularly pronounced in Arabic. As such, chrF++ is especially well-suited for DA--MSA translation tasks, where minor lexical or syntactic divergences often do not impair semantic equivalence.

Although embedding-based alternatives such as COMET \citep{rei-etal-2020-comet} and BERTScore \citep{zhang2019bertscore} offer the potential for improved semantic alignment, preliminary experiments that we conducted indicated suboptimal performance for dialectal Arabic. This limitation is likely due to the inadequate representational quality of Arabic embeddings in these models. While the LLM-as-a-judge paradigm has recently gained popularity in machine translation (MT) evaluation, where an LLM is prompted to assign scores reflecting the adequacy or correctness of a translation \citep{bavaresco-etal-2025-llms}, its use in this study was infeasible due to the prohibitive cost of large-scale prompting. Furthermore, such approaches offer limited interpretability, as the decision-making process behind the assigned scores remains opaque. Consequently, SacreBLEU and chrF++ were selected for their computational efficiency, interpretability \citep{papineni-etal-2002-bleu, Popovic2017chrF++}, and common use in previous low-resource machine translation research\citep{abdulmageed2024nadi, osact2024, mager-etal-2023-neural}.

These two metrics were employed to provide complementary perspectives: SacreBLEU serves as a precision-oriented measure of lexical fidelity, whereas chrF++ offers a more balanced assessment of precision and recall at the character level. Both metrics were computed for all model outputs. However, in all subsequent results analysis, greater emphasis was placed on chrF++ scores due to their superior suitability for Arabic’s inflectional and derivational morphology. This evaluation protocol was applied uniformly throughout all experiments to ensure methodological consistency and facilitate comparability.

\subsection{Training-Free Prompting Setup}\label{sec:training-free-prompting}
\subsubsection{Prompting Strategies}

This study examined four training-free prompting strategies for DA--MSA translation. All prompt templates used are included in Section ~\ref{sec:prompting-templates}.

\paragraph{\textbf{Zero-Shot Prompting}}
The Zero-Shot prompt leverages the pre-trained knowledge of the model to perform translation based solely on an explicit instruction (e.g. ``Translate this sentence from the Levantine Dialect to MSA''). Although this approach is efficient and often effective for high-resource language pairs~\citep{johnson2017google,brown2020language}, its performance in low-resource scenarios such as DA–MSA translation remains limited. Challenges such as informal orthography, code-switching, and semantic ambiguity frequently reduce translation accuracy. These limitations are consistent with previous findings that highlight the sensitivity of dialect-specific performance to the coverage and composition of the pre-training corpus~\citep{sajjad2023dialects,zhao2023exploring}.

\paragraph{\textbf{Zero-Shot Chain-of-Thought Prompting}}

Chain-of-Thought (CoT) prompting extends zero-shot prompting by instructing the model to generate intermediate \textit{reasoning} steps prior to producing the final translation \citep{wei2022chain}. Originally developed for arithmetic and logical inference tasks, CoT was adapted in this study to support dialectal translation. The rationale is that by articulating intermediate reasoning, the model can better resolve idiomatic expressions, lexical ambiguities, and context-dependent phrasing, thus improving translation accuracy.

\paragraph{\textbf{Few-Shot Prompting}}

Few-shot prompting improves the translation quality by including a limited number of DA--MSA \textit{example pairs} within the prompt. As demonstrated in previous low-resource research \citep{sajjad2023dialects}, this strategy enhances generalization by providing concrete examples of the translation task. For each dialect, we curated three representative examples corresponding to the following: conversational, descriptive, and idiomatic sentence types. These categories capture key linguistic features of dialectal Arabic and were selected to expose the model to register variation, structural diversity, and culturally embedded expressions.

\paragraph{\textbf{Ara-TEaR}}

To further enhance translation quality, we introduce \textit{Ara-TEaR}, an adaptation of the TEaR framework originally proposed by ~\cite{feng2024tear}. Ara-TEaR incorporates a structured three-stage prompting process: (1) initial translation generation, (2) self-evaluation of the translation against predefined quality criteria, and (3) output refinement based on self-assessment feedback.

\begin{figure}[H]
  \centering
  \includegraphics[width=1.0\linewidth]{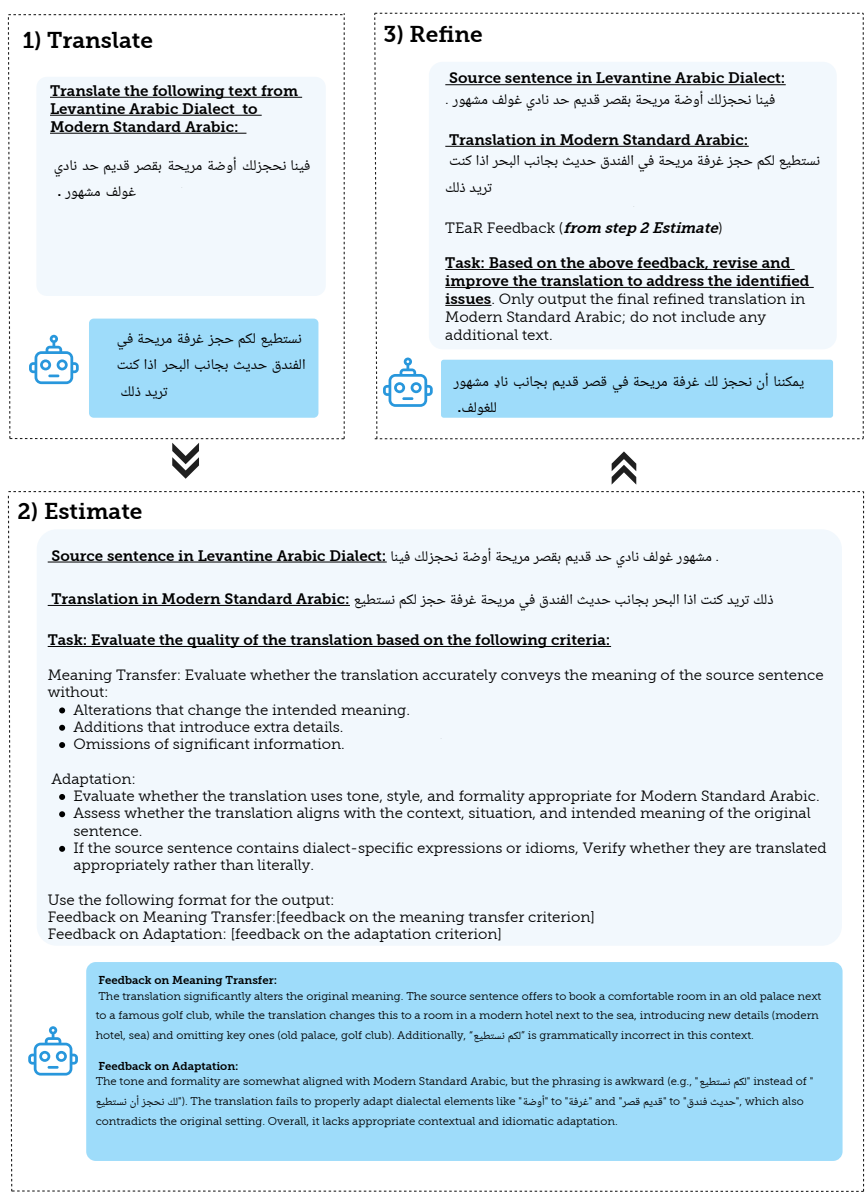}
  \caption{Illustration of the Ara-TEaR three-phase workflow tailored to address major DA-MSA translation errors.}
  \label{fig:ara-tear}
\end{figure}

While the original TEaR framework employs Multidimensional Quality Metrics (MQM) \citep{lommel2012mqm,lommel-etal-2024-multi}, Ara-TEaR is specifically \textit{tailored} to the DA--MSA context. It targets two frequent error types: \textit{Meaning Transfer} errors, in which the intended meaning is not adequately preserved, and \textit{Adaptation} errors, which occur when the style fails to shift appropriately from dialectal Arabic to formal MSA. These error types were identified through manual analysis of the outputs generated by three baseline models: GPT-3.5, Jais, and NLLB-200. A visual overview of the Ara-TEaR workflow is presented in Figure~\ref{fig:ara-tear}.

\subsubsection{Model and Dataset Selection}
In our model prompting experiments, we evaluate six LLMs to examine how prompting effectiveness varies across architectures and parameter scales. Previous research has shown that model performance is influenced by both architectural design~\citep{vaswani2017attention} and the composition of pre-training data~\citep{zhao2023exploring}. As shown in Table~\ref{tab:model_overview}, the selected models span a range of sizes and architectures, allowing a comprehensive analysis of their respective strengths and limitations in the DA--MSA translation task.

\begin{table}[htbp]
\centering
\begin{tabular}{lc}
\hline
\textbf{Model} & \textbf{Parameters} \\
\hline
GPT-3.5 & 175B \\
GPT-4o & Approximately 1.8T \citep{explodingtopics2025gpt} \\
LLaMA 3.3-70B-Instruct & 70B \\
LLaMA 3.1-405B & 405B \\
Gemma-27B & 27B \\
DeepSeek-V3 & 671B \\
\hline\hline
\end{tabular}
\caption{Overview of evaluated models and their parameter sizes, highlighting architectural diversity in the DA--MSA translation experiments}
\label{tab:model_overview}
\end{table}

Our model selection was intended to reflect a diverse spectrum of capabilities and design philosophies. GPT-3.5 was included as a widely adopted general-purpose baseline, while GPT-4o and DeepSeek-V3 represent recent advancements in multilingual LLMs. LaMA 3.3-70B-Instruct and LLaMA 3.1-405B facilitate analysis of instruction tuning and scaling in open-source systems. Gemma-27B was chosen for its computational efficiency and observed good performance during manual testing.

As no parameter updates were performed during prompting, training data was not required. Instead, evaluation relied solely on the \textit{Small-Test} test set, derived from the MADAR corpus, comprising 600 examples. While this test set is smaller than our \textit{Large-Test} dataset (1,200 examples), which is used in the model fine-tuning experiment, \textit{Small-Test} dataset's size was also chosen with practical considerations in mind. Given the prohibitive token-based usage costs associated with API access to these models, \textit{Small-Test} was designed to be small enough to allow us to complete all our experiments, which involve extensive evaluation across different models and prompting strategies. This setup supported our systematic and resource-conscious evaluation across the different model variants.


\subsubsection{Implementation Details}

The prompting experiments were conducted via API access to each of the selected models. Specifically, the OpenAI API was utilized for GPT-3.5 and GPT-4o, the Anthropic API for Claude 3.5 Sonnet, and the LLAMA API for the open-source models: LLaMA 3.3-70B-Instruct, LLaMA 3.1-405B, Gemma-27B, and DeepSeek-V3.

To ensure consistency across models and experiments, key generation parameters were standardized. The maximum number of tokens (\texttt{max\_tokens}) was set to 512 to avoid truncation of output. The temperature was fixed at 0.3, balancing between creativity and consistency, while \texttt{top\_p} was set to 0.9 to allow for controlled lexical variation. These settings were selected to promote the generation of fluent, formal MSA translations while preserving the linguistic nuances of the source dialectal input.

\subsection{Model Fine-Tuning Setup}\label{sec:model_fine_tune_setup}
\subsubsection{Model and Dataset Selection}

The Gemma-2-9B model was selected as the primary architecture for the majority of fine-tuning experiments due to its balance between computational efficiency and translation quality. In addition to its effectiveness in multilingual tasks, particularly for non-English language pairs, Gemma-2-9B demonstrates strong translation capabilities across many languages~\citep{cui2025multilingual}, making it suitable for DA--MSA translation. With 9 billion parameters, it provides sufficient capacity to capture dialectal variation while remaining computationally feasible for fine-tuning. Additional features of this model, such as native support for 4-bit quantization and LoRA, further reduced memory requirements, enabling efficient training under low-resource conditions.

Fine-tuning experiments utilized the three dataset compositions: \textit{Minimal-Gold} , \textit{Extended-Gold}, and \textit{Extended-Gold-Silver}. All data were formatted using an Alpaca-style instruction template~\citep{taori2023alpaca}, adapted for DA--MSA translation:

\begin{tcolorbox}[colframe=black, colback=verylightgray, arc=5mm, boxrule=0.8pt]
Translate the below text from Dialectal Arabic to Modern Standard Arabic:

\#\#\# DA text: \{\}

\#\#\# MSA translation: \{\}
\end{tcolorbox}

This format capitalizes on the instruction-following capabilities of modern LLMs while omitting explicit dialect labels, thus encouraging implicit dialect identification. To mitigate any training data biases, all training datasets were randomly shuffled.

Evaluation primarily relied on the \textit{Large-Test} test set, which comprises 1,200 examples (400 per dialect) drawn from both MADAR (70\%) and Dial2MSA-Verified (30\%). The heavier weighting toward MADAR reflects its greater reliability due to professional translation, while the inclusion of Dial2MSA-Verified data introduces additional lexical and stylistic diversity from social media contexts. During initial hyperparameter optimization, the smaller \textit{Small-Test} set (600 examples) was employed as a development set to reduce evaluation costs. The size of our dataset aligns with common DA-MSA shared tasks, such as NADI-2024 \citep{abdulmageed2024nadi}, where the test dataset size was 400 examples per dialect, and OSACT-2024~\citep{osact2024}, where the dataset sizes for the Egyptian, Levantine, and Gulf dialects were 314, 568, and 586 examples, respectively.

\subsubsection{Fine-Tuning Configuration}

The experiments were conducted using A100 GPUs on Google Colab Pro. LoRA was applied with a rank and an alpha parameter of 16. The key training settings included a batch size of 16 and gradient accumulation steps of 2. The AdamW optimizer was employed in 8-bit mode, with a weight decay coefficient of 0.01. A linear learning rate scheduler was utilized, and mixed precision training (FP16/BF16) was adopted to improve computational efficiency.

To ensure reproducibility, a fixed random seed (3407) was used in conjunction with deterministic data splits. The learning rate and number of training epochs were optimized. Input sequences were capped at a maximum length of 2,048 tokens, and end-of-sequence (EOS) tokens were appended to prevent infinite text generation during inference.

\subsubsection{Experimental Framework} \label{sec:exp_framework}
A series of ablation experiments (8 in total) was conducted to systematically assess factors influencing the performance of our models, including model architecture, data strategy, and training hyperparameters. Beginning with a baseline configuration, each subsequent experiment introduced a controlled modification to isolate the effect of a single variable. This structured approach facilitated reproducible comparisons and yielded targeted insights for optimizing DA--MSA translation in low-resource settings. The 8 ablation experiments are named as below and will be explained in details.
\begin{itemize}
    \item \textsc{Base}: {Baseline Performance (v0)}
    \item \textsc{HypperParam}: {Hyperparameter Optimization (v1)}
    \item \textsc{Join-vs-Spec}:  {Joint vs.\ Dialect-Specific Models (v2)}
    \item \textsc{DataScale}: {Data Scaling Effect (v3)}
    \item \textsc{Augmentation}: {Data Augmentation Effect (v4)} 
    \item \textsc{Architecture}: {Model Architecture Comparison (v5)}\item \textsc{Quantiz}: {Quantization Impact (v6)}
    \item \textsc{ActiveLearn}: {Active Learning Approach (v7)}
\end{itemize}

\paragraph{\textsc{Base}: \textbf{Baseline Performance (v0)}}

The primary objective of this experiment was to establish a baseline for DA--MSA translation performance using an untuned model. This baseline served as a reference point for evaluating the impact of subsequent fine-tuning experiments. The evaluation focused on the Gemma-2-9B model in two configurations: the base model (without instruction tuning) and an instruction-tuned variant. Both configurations were evaluated under a zero-shot inference setting, without any fine-tuning. All evaluations were conducted on the designated test set, following the setup described earlier.

\paragraph{\textsc{HypperParam}: \textbf{Hyperparameter Optimization (v1)}}

This experiment aimed to identify optimal values for the learning rate and the number of training epochs to balance model performance and training stability. These hyperparameters were prioritized due to their substantial impact on convergence behavior and overfitting. The \textit{Minimal-Gold}  dataset (MADAR: 43,000 samples) was selected for its sufficient size to support meaningful training while avoiding the complexity and noise associated with larger datasets. The Gemma-2-9B model was employed with consistent architectural and training configurations, while systematically varying the learning rate (1e-4 vs.\ 5e-5) and the number of training epochs (1 vs.\ 3).

\paragraph{\textsc{Join-vs-Spec}:  \textbf{Joint vs.\ Dialect-Specific Models (v2)}}

This experiment investigated whether a joint multi-dialect model could outperform dialect-specific models by leveraging shared linguistic features across dialects. The Gemma-2-9B model was fine-tuned under two configurations:

\begin{enumerate}
  \item \textit{Joint Model:} Trained on the complete \textit{Minimal-Gold}  dataset, covering all three dialects (LEV, GLF, EGY).
  \item \textit{EGY-Specific Model:} Trained exclusively on the Egyptian subset of the \textit{Minimal-Gold}  dataset.
\end{enumerate}

\paragraph{\textsc{DataScale}: \textbf{Data Scaling Effect (v3)}}

This experiment evaluated the impact of increasing training data volume and introducing domain diversity on translation performance. The Gemma-2-9B model was fine-tuned using the \textit{Extended-Gold}  dataset (MADAR + Dial2MSA-Verified), expanding the training set from 43,000 to 62,775 examples. All training configurations were kept consistent with those employed in the previous experiment.

\paragraph{\textsc{Augmentation}: \textbf{Data Augmentation Effect (v4)} }

This experiment aimed to evaluate whether incorporating synthetic data generated through MSA paraphrasing could improve DA--MSA translation quality. The \textit{Extended-Gold-Silver}  training dataset, combined the \textit{Extended-Gold}  dataset (comprising 62,775 samples) with two additional components: 43,000 paraphrased MSA sentences generated using GPT-4o, and 5,000 same-source-target MSA--MSA pairs extracted from the UFAL corpus. This resulted in a total of 110,775 training samples. Two model variants, Gemma-9B and Gemma-27B, were fine-tuned on this dataset to investigate whether model size influences the effectiveness of data augmentation. GPT-4o was chosen for paraphrase generation due to its strong zero-shot performance and relatively low cost.

\paragraph{\textsc{Architecture}: \textbf{Model Architecture Comparison (v5)}}

This experiment evaluated DA--MSA translation across various model architectures and sizes. Its objectives are to determine whether smaller fine-tuned models could match the performance of larger zero-shot models, identify architecture-specific strengths, and provide practical insights for deployment in resource-constrained settings. Models were selected based on their architectural diversity, parameter size, and multilingual capabilities. The \textit{Extended-Gold}  dataset (MADAR + Dial2MSA-Verified ), identified as the best-performing configuration in the \textsc{DataScale} experiment, was used for training. All models were fine-tuned using identical hyperparameters (learning rate = 5e-5 and one epoch). Table~\ref{tab:model_comparison} presents the fine-tuned and zero-shot prompted models evaluated in this experiment.

\begin{table}[htbp]
\centering
\begin{tabular}{lcc}
\hline
\textbf{Fine-Tuned Models} & & \textbf{Zero-Shot Prompted Models} \\
\hline
LLaMA-3.2-3B-Instruct & & Deepseek V3 \\
LLaMA-3.1-8B & & LLaMA-3.3-70B \\
Gemma2-9B & & Jais 30B \\
Gemma2-27B & & GPT-3.5 \\
 & & GPT-4o \\
\hline\hline
\end{tabular}
\caption{Fine-tuned and zero-shot models evaluated in the \textsc{Architecture} experiment}
\label{tab:model_comparison}
\end{table}

\paragraph{\textsc{Quantiz}: \textbf{Quantization Impact (v6)}}

This experiment assessed the trade-offs between translation quality and computational efficiency resulting from applying 4-bit quantization. The objective is to validate the use of quantization in our prior experiments and examine its viability for real-world deployment. The LLaMA-3.2-3B-Instruct model was fine-tuned on the \textit{Extended-Gold}  dataset using two precision settings: full-precision (16-bit), which served as the baseline, and 4-bit quantization, which was used to explore potential resource savings and quality retention. This model was selected due to its smaller size, which enabled fine-tuning under both configurations given the available hardware resources.

\paragraph{\textsc{ActiveLearn}: \textbf{Active Learning Approach (v7)}}
This experiment investigated the effect of selective fine-tuning using linguistically challenging examples to determine whether emphasizing complex sentences could improve generalization in dialect-to-MSA translation. The Gemma-9B model, previously fine-tuned on the \textit{Extended-Gold}  dataset, was further fine-tuned using a subset consisting of the longest 30\% of sentences from the \textit{Extended-Gold}  dataset. This subset was selected based on the assumption that longer sentences are more likely to exhibit greater syntactic and semantic complexity.

\section{Experimental Results} 
\subsection{Evaluation of the Training-free prompting approach for DA--MSA translation}\label{sec:training-free-prompting-eval}

\subsubsection{Models performance across prompting strategies}
This section analyses the translation performance of models in our first approach to DA-MSA translation. The reported scores are averaged across the three dialects represented in the test set. Detailed per-dialect results are available in Section~\ref{sec:Detailed-Results}. Table \ref{tab:model_performance} summarizes the average chrF++ scores for each model and prompting strategy.

To determine if there were statistically significant differences in performance among the four prompting strategies for each model, we conducted a repeated measures Analysis of Variance (ANOVA). In our design, the four different prompting strategies represent the conditions under which each model was repeatedly tested.

The two primary assumptions for ANOVA, normality and sphericity, were verified. We assessed the normality of the data distribution using the Shapiro-Wilk test, setting a significance level of $0.05$ that must be exceeded to assume normality. To complement this analytical approach, we also visually inspected histograms and Quantile-Quantile (QQ) plots. Sphericity was evaluated using Mauchly's Test of Sphericity, with the assumption being met if $p \le 0.05$.

When the ANOVA indicated a statistically significant overall effect, we performed Bonferroni post-hoc tests (with $0.05$ threshold for the $p$-value) to identify which specific pairs of prompting strategies differed significantly from one another. The results of these pairwise comparisons are presented in Table~\ref{tab:strategy_stat_sign_comparison}.

\begin{table}[htbp]
\centering
\begin{tabular}{lcccc}
\hline
Model & 0‑Shot & 0‑Shot CoT & Ara-TEaR & Few‑Shot \\
\hline
GPT‑3.5                  & 38.51 & 36.74 & 35.32 & 35.93 \\
GPT‑4o                   & 42.18 & 42.01 & 41.25 & 42.67 \\
LLaMA 3.3‑70B‑Instruct   & 34.94 & 34.09 & 35.27 & 35.4 \\
LLaMA 3.1‑405B           & 37.01 & 36.47 & 35.89 & 37.26 \\
Gemma‑27B                & 35.52 & 36.04 & 35.66 & 36.09 \\
DeepSeek‑V3              & 40.22 & 39.72 & 38.84 & 40.6 \\
\hline\hline
\end{tabular}
\caption[Performance Comparison Across Models and Prompting Strategies]
{Average chrF++ scores comparing models performance with zero-shot, CoT, Ara-TEaR and Few-shot prompting strategies for DA-MSA translation}
\label{tab:model_performance}
 \end{table}

\begin{table}[htbp]
\centering
\begin{tabular}{lccc}
\hline
\textbf{Model} & \textbf{Strategy A} & \textbf{Strategy B} & \textbf{p-value} \\
\hline
GPT‑3.5 & 0‑Shot & 0‑Shot CoT & \textless0.001 \\
GPT‑3.5 & 0‑Shot & Ara-TEaR & \textless0.001 \\
GPT‑3.5 & 0‑Shot & Few‑Shot & \textless0.001 \\
GPT‑3.5 & 0‑Shot CoT & Ara-TEaR & 0.046 \\
GPT‑4o & Few‑Shot & Ara-TEaR & 0.026 \\
LLaMA 3.3‑70B‑Instruct & Few‑Shot & 0‑Shot CoT & 0.046 \\
LLaMA 3.1‑405B & Few‑Shot & Ara-TEaR & 0.044 \\
DeepSeek‑V3 & Few‑Shot & Ara-TEaR & 0.033 \\
\hline\hline
\end{tabular}
\caption{Statistically significant results where Strategy A achieves a statistically more significant score than Strategy B}
\label{tab:strategy_stat_sign_comparison}
\end{table}

\paragraph{\textbf{Effectiveness of Prompting Strategies}}
Few-shot prompting emerges as the most effective strategy across the majority of tested models, including GPT-4o, LLaMA 3.3‑70B‑Instruct, LLaMA 3.1‑405B, Gemma‑27B, and DeepSeek-V3. The consistent superiority of this method is likely due to the use of concrete translation examples tailored to each dialect, which enhances the models’ understanding of dialectal structures. By providing diverse and context-rich examples, few-shot prompting enables models with strong instruction-following capabilities to generalize more effectively and capture subtle linguistic nuances, hence achieving higher chrF++ scores.

Zero-shot prompting serves as a useful baseline for understanding the translation abilities of each model. Interestingly, GPT-3.5 achieves its highest performance using this simplest strategy, outperforming more sophisticated prompting techniques. This outcome suggests that GPT-3.5 may have been exposed to substantial Arabic or multilingual content during pretraining, rendering additional guidance unnecessary or even counterproductive.

The use of Chain-of-Thought (CoT) reasoning in a zero-shot setting yields only marginal improvements, with a modest benefit observed for Gemma-27B. For the majority of models, CoT does not significantly enhance performance, implying that explicitly generating intermediate reasoning steps might not consistently lead to better chrF++ scores, or perhaps the nature of the reasoning generated by the models is not always conducive to higher translation quality in this specific low-resource context. This contrasts with CoT's effectiveness in arithmetic and logical inference tasks.

Finally, the Ara-TEaR method fails to demonstrate any statistically significant performance gains across models. In fact, higher-performing models tend to show a consistent decline when this approach is applied. This suggests that the self-evaluation and refinement mechanisms in Ara-TEaR, as currently implemented, may not be effective in identifying or correcting translation errors in the DA-MSA task.

\paragraph{\textbf{Model Architecture and Scaling Effects}}
GPT-4o consistently outperforms all other models across the evaluated prompting strategies, achieving the highest chrF++ scores overall. Its performance ranges from 41.25 in the Ara-TEaR setting to 42.67 in the Few-Shot setting. The relatively narrow performance range across strategies suggests that GPT-4o possesses robust internal representations, making it less sensitive to variations in prompt structure. This indicates a strong underlying capacity for DA-MSA translation that is not heavily reliant on specific prompting techniques.

DeepSeek-V3 emerges as the second-best performer in most prompting strategies. However, its performance declines more noticeably in Ara-TEaR settings, indicating that it may be less effective at leveraging self-evaluation mechanisms in this translation task.

GPT-3.5 demonstrates competitive performance in the simpler prompting strategies like Zero-Shot. However, the model shows a clear decline in translation quality when more complex prompting strategies, like Ara-TEaR, are applied. This drop suggests that GPT-3.5 may not effectively utilize refinement-driven prompts, possibly due to limitations in model training.

Model scaling within the same family appears to correlate positively with performance. For example, LLaMA 3.1‑405B consistently outperforms the smaller LLaMA 3.3‑70B‑Instruct across all prompting strategies. However, this trend does not hold universally across model families. Notably, Gemma‑27B outperforms LLaMA 3.3‑70B‑Instruct in all strategies, indicating that factors beyond size, such as model architecture and pre-training data, are critical. These results underscore the need to consider both scaling and design choices when evaluating model performance in low-resource translation tasks.
\newline

\begin{figure}[t]
    \centering
    \includegraphics[width=0.9\linewidth]{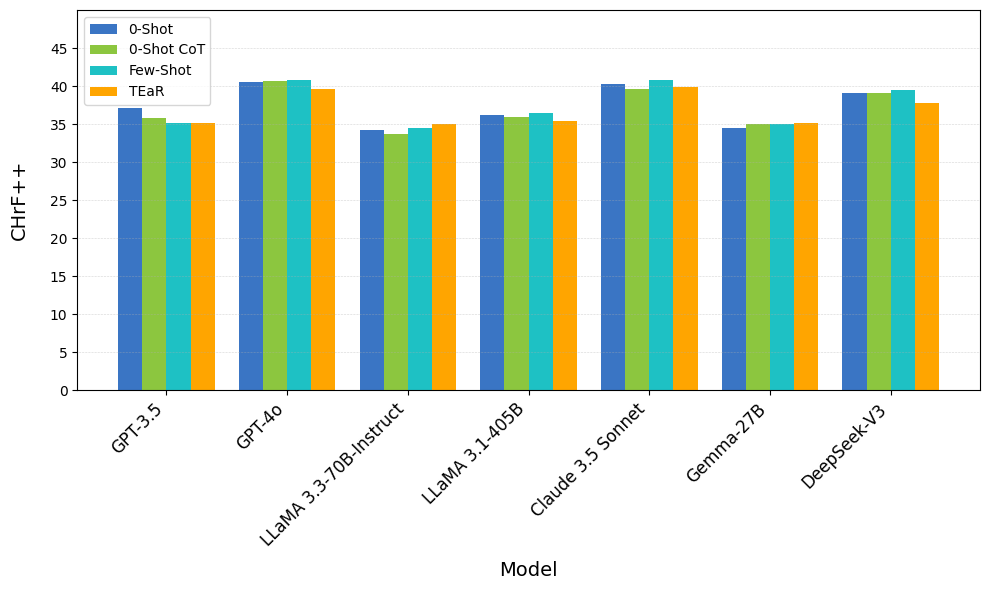}
    \caption{Comparison of chrF++ scores across models using Zero-Shot, Zero-Shot CoT, Few-Shot, and Ara-TEaR strategies for DA-MSA translation}
    \label{fig:performance_comparison}
\end{figure}

Figure~\ref{fig:performance_comparison} provides a visual representation of the performance comparison across different models and prompting strategies, illustrating the key patterns discussed above.

\subsubsection{Dialect-identification impact on prompting performance}

An experiment involving GPT-4o and LLaMA 3.3-70B-Instruct was conducted to evaluate the effect of explicitly specifying the source dialect in zero-shot prompts. Two prompt formulations were compared: one explicitly named the source dialect (e.g. ``Levantine''), while the other used the generic term ``Dialectal Arabic''. The results indicate that explicit dialect identification improves translation quality for both models.

GPT-4o exhibited a moderate improvement in chrF++ scores, increasing from 39.08 to 40.52. This gain reflects the model’s strong multilingual capabilities, likely stemming from extensive and diverse pre-training. In contrast, LLaMA 3.3-70B-Instruct demonstrated a more substantial improvement, with chrF++ scores rising from 30.93 to 34.26. This suggests that LLaMA 3.3-70B-Instruct has limited capacity for implicit dialect identification and relies more heavily on explicit prompt structure to guide its translation decisions.

The observed disparity highlights key architectural and training differences between the models. While GPT-4o is better equipped to handle vague or underspecified prompts due to its broader multilingual exposure, LLaMA benefits significantly from structured and precise input. These findings underscore the critical role of prompt design in optimizing model performance, particularly for systems with narrower pre-training coverage or smaller parameter counts.

\subsubsection{Cross-dialect performance variability}

An analysis of per-dialect performance (presented in Section ~\ref{sec:Detailed-Results}) reveals that Egyptian Arabic consistently yields the highest chrF++ scores across nearly all models in the zero-shot setting, followed by Gulf Arabic and then Levantine. This outcome contrasts with established linguistic research, which identifies Levantine Arabic as morphologically and syntactically more similar to MSA than many other dialects \citep{abukwaik2019lexical}. The observed disparity is likely attributable to the disproportionate representation of Egyptian Arabic in pre-training data, reflecting its widespread use in media and online content.

These findings suggest that the amount of dialectal exposure during model pre-training exerts a stronger influence on translation performance than linguistic proximity to MSA. Consequently, the results underscore the limitations of relying on linguistic similarity alone when designing prompts or evaluating model performance. Instead, they emphasize the importance of dialect-aware prompting strategies that account for data availability and representation in model pre-training.

\subsubsection{Computational Efficiency Analysis}
\begin{figure}[t]
    \centering
    \includegraphics[width=0.9\linewidth]{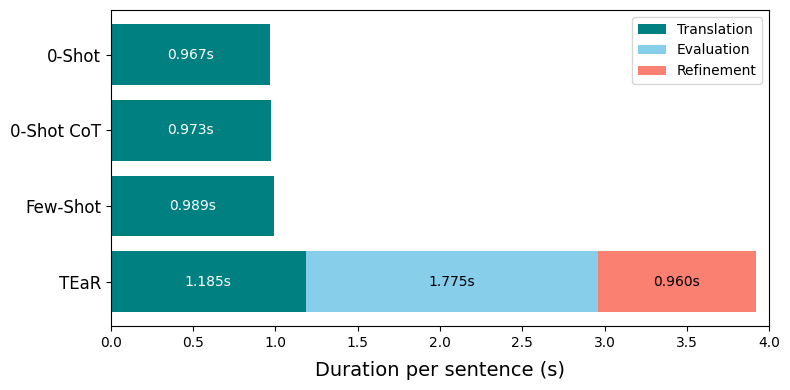}
    \caption{Comparison of average response times for zero-shot, chain-of-thought, few-shot, and Ara-TEaR strategies during DA-MSA translation}
    \label{fig:cost_comparison}
\end{figure}

Figure~\ref{fig:cost_comparison} presents the average time required to translate 200 Levantine sentences to MSA using Gemma2-9B via the LLAMA API, comparing the model's response time for each prompting strategy. Zero-shot (0.967s) and zero-shot CoT (0.973s) show minimal difference, indicating negligible overhead from reasoning prompts. Few-shot (0.989s) is slightly slower due to example processing.

Ara-TEaR is significantly more demanding (3.92s), taking nearly four times more time than simpler methods. Its evaluation phase (1.775s) is the main bottleneck, accounting for ~45\% of total time, while translation (1.185s) and refinement (0.960s) are less intensive.

Token usage follows similar patterns, with Ara-TEaR requiring substantially more tokens due to its three-phase process. This translates to higher operational costs and computational demands, limiting its feasibility in practical deployment scenarios, particularly for real-time applications or resource-constrained environments.

\subsection{Evaluation of the Resource-efficient fine-tuning approach for DA--MSA translation}\label{sec:fine-tuning-eval}

This section presents the results for the experiments outlined in Section ~\ref{sec:exp_framework}.

\subsubsection{\textsc{Base}: Baseline Performance (v0)}
The base model (Gemma-2-9B) yielded a near-zero chrF++ score of 1.48 on the test set. Manual review indicated that the model outputs failed to adhere to the translation instructions, prompting our decision to instead evaluate the instruction-tuned variant (Gemma-2-9B-IT) which demonstrated moderate performance across the three dialects, with chrF++ scores as follows:
\begin{itemize}
    \item Levantine (LEV): 29.91
    \item Gulf (GLF): 33.59
    \item Egyptian (EGY): 30.83
    \item Average score acorss dialects: 31.44
\end{itemize}
Although the model was not explicitly trained on dialect-specific data, it produced relatively consistent results across dialects. The narrow score range ($\leq 3.8$ chrF++) suggests minimal architectural bias toward any particular dialect, validating the model's suitability as a baseline for our fine-tuning experiments.

\subsubsection{\textsc{HypperParam}: Hyperparameter Optimization (v1)}
\paragraph{\textbf{HypperParam-LR: Learning Rate Comparison}} 
Experiment Results (Validation Set):
\begin{itemize}
    \item Learning rate (LR) = 1e-4: chrF++ score of 43.72
    \item Learning rate (LR) = 5e-5: chrF++ score of 43.03
\end{itemize}

\begin{figure}[t]
    \centering
    \includegraphics[width=0.9\linewidth]{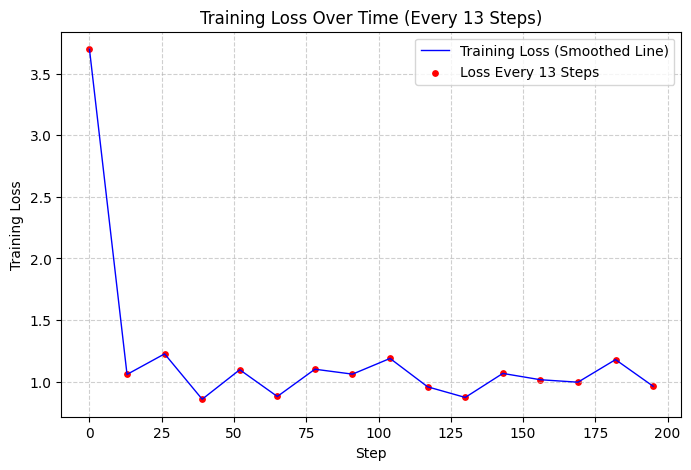}
    \caption{Training Loss Curve During Fine-Tuning}
    \label{fig:curve}
\end{figure}

Figure~\ref{fig:curve} illustrates the training loss curve for this experiment, using a smoothed line to emphasize the overall trend and overlook short-term fluctuations. The loss curves for both learning rates were nearly identical. Although the higher learning rate of 1e-4 produced a slightly higher score, the learning rate of 5e-5 was selected due to its likely greater stability and lower risk of oscillations, which is especially important when scaling to larger datasets or longer training epochs in subsequent experiments.

\paragraph{\textbf{HypperParam-Epoch: Epoch Comparison}} 

Experiment Results (Validation Set):
\begin{itemize}
    \item 1 Epoch: chrF++ score of 45.64
    \item 3 Epochs: chrF++ score of 40.59
\end{itemize}
Training for a single epoch outperformed training for three epochs. This is likely because shorter training reduced the risk of overfitting. Extended training appeared to cause model divergence, which negatively affected generalization to unseen data.
\newline

The best configuration achieved a chrF++ score of 43.44, representing a substantial improvement over the baseline score of 31.44 established in the \textsc{Base} Experiment. Model fine-tuning and hyperparameter-tuning resulted in a 38\% improvement in chrF++ over the baseline. Therefore, all subsequent experiments use a learning rate of 5e-5 and a single training epoch.

\subsubsection{\textsc{Join-vs-Spec}:  Joint vs.\ Dialect-Specific Models (v2)}
\begin{figure}[t]
    \centering
    \includegraphics[width=0.9\linewidth]{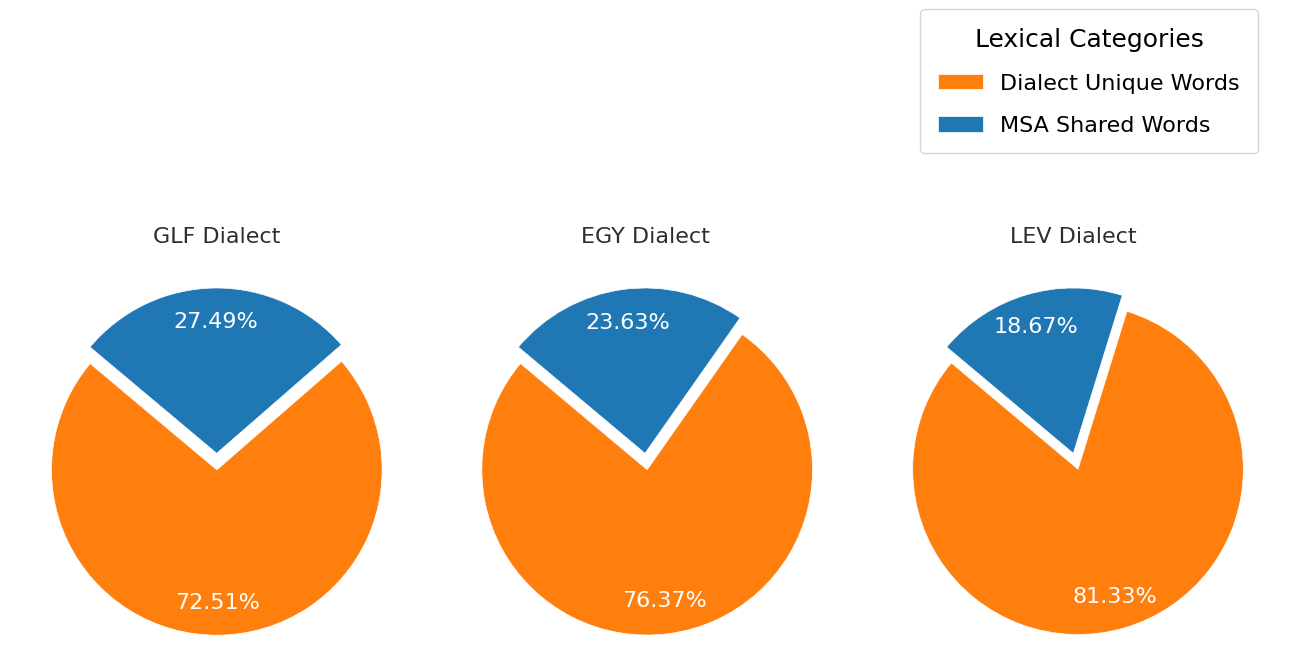}
    \caption{Proportion of shared vocabulary between Modern Standard Arabic and Levantine, Gulf, and Egyptian dialects in the training dataset}
    \label{fig:msa-share}
\end{figure}

Experiment Results for the EGY Dialect (Test Set):
\begin{itemize}
    \item Joint Model: chrF++ score of 42.72
    \item EGY-Specific Model: chrF++ score of 38.62
\end{itemize}
The joint model outperformed the dialect-specific model by 10.6\%, highlighting the benefits of cross-dialect training. As shown in Figure~\ref{fig:msa-share}, there is substantial lexical overlap between MSA and the dialects: 18.7\% for Levantine (LEV), 23.6\% for Gulf (GLF), and 27.5\% for Egyptian (EGY). This overlap, along with shared grammatical structures, enables better generalization in joint models. These results provide strong empirical support for employing a joint modeling strategy in subsequent experiments.

\subsubsection{\textsc{DataScale}: Data Scaling Effect (v3)}
Experiment Results (Test Set):
\begin{itemize}
    \item Model trained on \textit{Extended-Gold}  dataset: chrF++ score of 49.88
    \item Model trained on the \textit{Minimal-Gold}  dataset (from \textsc{HypperParam} experiment): chrF++ score of 43.44
\end{itemize}

Training on the \textit{Extended-Gold}  dataset yielded a 14.8\% improvement in chrF++ over the model trained on the \textit{Minimal-Gold}  dataset. A significant contributor to this gain was the inclusion of Dial2MSA-Verified dataset (which has a social media domain). Hence, enhancing the model's capacity to handle informal expressions and domain-specific linguistic patterns. These results underscore the value of both enlarging the training corpus and diversifying its domain.

\subsubsection{\textsc{Augmentation}: Data Augmentation Effect (v4)}
Experiment Results:
\begin{itemize}
    \item Gemma-9B: chrF++ score of 46.92 (compared to 49.88 without data augmentation)
    \item Gemma-27B: chrF++ score of 16.79, indicating severe degradation
\end{itemize}

\paragraph{\textbf{Data Quality vs. Noise}} 

Despite manual validation confirming the linguistic accuracy of the synthetic paraphrases, the augmented data has likely introduced redundant lexical patterns. This redundancy led to overfitting in the Gemma-9B model, reducing its cross-dialect generalization.

\paragraph{\textbf{Model Size Sensitivity}} 

The pronounced drop in performance for Gemma-27B suggests that larger models are particularly sensitive to noisy or repetitive data. With greater capacity, these models are more prone to memorization, making them less robust without carefully curated training dataset.
\newline

Due to computational constraints, this study could not investigate optimal synthetic-to-gold data ratios (e.g., 1:3), which may mitigate overfitting. Overall, the results emphasize that even high-quality synthetic data must be rigorously filtered, especially when training larger models, to avoid detrimental effects on performance. While our synthetic dataset did not improve the performance, we made it publicly accessible to allow for further investigation and potential use in other Arabic NLP tasks.

\subsubsection{\textsc{Architecture}: Model Architecture Comparison (v5)}
\begin{table}[htbp]
\centering
\begin{tabular}{lc}
\hline
\textbf{Model} & \textbf{chrF++} \\
\hline
\multicolumn{2}{l}{\textit{         Fine-Tuned Models Performance}} \\
Gemma2-9B & 49.88 \\
Gemma2-27B & 48.49 \\
Llama-3.1-8B & 46.25 \\
Llama-3.2-3B-Instruct & 37.37 \\
\hline
\multicolumn{2}{l}{\textit{         Zero-Shot Models Performance}} \\
GPT-4o & 44.58 \\
DeepSeek V3 & 41.05 \\
GPT-3.5 & 38.43 \\
LLaMA3-70B & 35.89 \\
Jais-30B & 31.22 \\
\hline\hline
\end{tabular}
\caption{Performance of fine-tuned and zero-shot models}
\label{tab:model_performance}
\end{table}

The results, as shown in Table \ref{tab:model_performance}, indicate that task-specific fine-tuning often outweigh raw parameter count. Gemma2-9B outperformed its larger counterpart, Gemma2-27B, by 1.39 chrF++, pointing to potential diminishing returns from scale alone without improving the training strategy and training dataset.

\paragraph{\textbf{Gemma vs. LLaMA Performance}} 
Gemma models consistently outperformed their LLaMA counterparts. For example, Gemma2-9B exceeded Llama-3.1-8B by 3.63 chrF++, despite similar model sizes. This likely reflects the advantage of Gemma's robust multilingual pretraining, which enhances its ability to handle dialectal variation.

\paragraph{\textbf{Fine-Tuning vs. Zero-Shot Performance}} 
Smaller fine-tuned models outperformed larger zero-shot models. Llama-3.2-3B-Instruct (37.37) outperformed the zero-shot Jais-30B (31.22) by 6.15 chrF++.

\begin{figure}[t]
    \centering
    \includegraphics[width=0.9\linewidth]{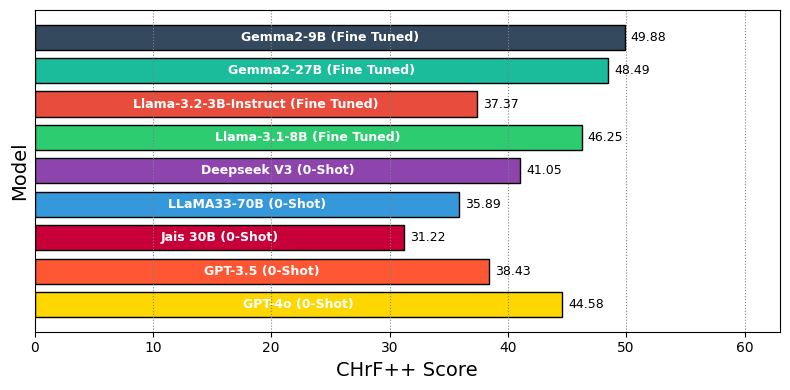}
    \caption{Performance comparison between fine-tuned and zero-shot models on the \textit{Large-Test} dataset}
        \label{fig:models}
\end{figure}

\paragraph{\textbf{Gemma2-9B vs. GPT-4o Performance}} 
Gemma2-9B emerged as the best performing model, combining strong performance (49.88 chrF++) with computational efficiency. Its 5.3 point advantage over GPT-4o underscores the potential of medium-sized fine-tuned models to surpass even state-of-the-art general-purpose models in specialized tasks like dialectal translation.
\newline
\newline
In summary, the findings underscore the importance of the model architecture choice and task-specific fine-tuning over sheer size. Gemma2-9B presents a compelling balance of accuracy and efficiency, making it well-suited for real-world deployment, particularly in resource-constrained environments. Figure~\ref{fig:models} provides an alternative visual presentation of the results.

\subsubsection{\textsc{Quantiz}: Quantization Impact (v6)}
Experiment Results:
\begin{itemize}
    \item Full-Precision Model: chrF++ score of 37.71
    \item 4-bit Quantized Model: chrF++ score of 37.37
\end{itemize}

\paragraph{\textbf{Performance Retention}} 
The 4-bit quantized model exhibited only a 0.9\% decrease in chrF++ score, a negligible reduction that confirms the viability of quantization while preserving translation quality. This supports the validity of using 4-bit quantization in earlier experiments, as the impact on performance remains within acceptable margins for practical use.
\paragraph{\textbf{Resource Efficiency Gains}} 
\begin{itemize}
    \item \textit{Memory Usage:} VRAM consumption dropped by 60\%, from 12.1GB to 4.8GB, enabling deployment on consumer-grade GPUs.
    \item \textit{Inference Speed:} Inference time improved by 22\%, decreasing from 1.8 to 1.4 seconds per batch, primarily due to the efficiency of low-bit matrix operations.
\end{itemize}

\paragraph{\textbf{Practical Implications}}
The modest performance trade-off, combined with significant reductions in memory usage and inference latency, highlights the effectiveness of 4-bit quantization for real-world deployment. These results underscore the practicality of using quantization techniques to enable scalable and accessible dialectal machine translation in resource-constrained settings.

\subsubsection{\textsc{ActiveLearn}: Active Learning Approach (v7)}
Experiment Results:
\begin{itemize}
    \item Baseline (fine-tuned on \textit{Extended-Gold}  dataset): chrF++ score of 49.88
    \item After Active Learning: chrF++ score of 49.15
\end{itemize}

Active learning led to a 1.5\% reduction in chrF++ score relative to the baseline, indicating that the sentence selection strategy was suboptimal. While longer sentences were assumed to offer richer syntactic structures, we suspect two key limitations after performing manual analysis of the sentences selected for active learning:

\begin{itemize}
    \item \textit{Redundant Context:} Many selected sentences included repetitive clauses or extraneous details. This redundancy introduced noise, causing the model to overfit on non-informative patterns rather than learning broadly applicable features.
    \item \textit{Limited Lexical Diversity:} The selected sentences exhibited low vocabulary variation, which restricted the model's ability to generalize across dialects and reduced the effectiveness of the fine-tuning process.
\end{itemize}

These findings suggest that effective active learning strategies should incorporate selection criteria that emphasize both syntactic richness and lexical diversity to better support generalization and improve translation quality.

\subsection{Comparative evaluation between model fine-tuning and training-free prompting for DA--MSA translation}

\begin{figure}[t]
    \centering
    \includegraphics[width=0.9\linewidth]{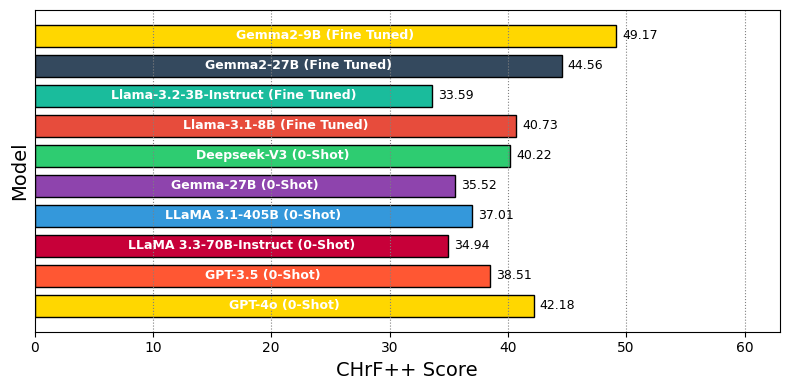}
    \caption{Comparison of chrF++ scores across different fine-tuned and prompted models on the \textit{Small-Test} dataset}
    \label{fig:prompting-vs-ft}
\end{figure}

While the evaluation of training-free prompting techniques in Section~\ref{sec:training-free-prompting-eval} was conducted on the \textit{Small-Test} dataset due to the prohibitive cost of model prompting, the larger (\textit{Large-Test}) dataset was used for evaluating fine-tuned models in Section~\ref{sec:fine-tuning-eval}. To enable direct comparison between the two approaches, in this section we evaluate all developed solutions (both fine-tuned and prompted models) on the \textit{Small-Test} dataset. Figure~\ref{fig:prompting-vs-ft} presents a summary of the results.

The results demonstrate that fine-tuning generally outperforms zero-shot prompting, with Gemma2-9B achieving the highest chrF++ score (49.17), surpassing even GPT-4o (42.18) despite its vastly smaller size. Gemma2-27B also performs strongly (44.56) but trails behind the 9B variant, suggesting diminishing returns from scaling under limited or noisy data. Fine-tuned LLaMA models achieve moderate gains, while prompted models cluster in the mid-30s to low-40s range. Overall, the findings highlight that \textit{targeted fine-tuning of mid-sized models provides the best balance of accuracy and efficiency for DA–MSA translation}.

\section{Conclusions}

This work focused on evaluating and advancing DA-MSA machine translation through two approaches: training-free prompting and resource-efficient fine-tuning.

The training-free prompting experiments demonstrated that few-shot prompting consistently yielded strong performance across various LLMs, with GPT-4o achieving the highest chrF++ scores. Our work also revealed that explicitly identifying the source dialect in prompts notably improved translation quality, particularly for models like LLAMA 3.3-70B-Instruct. Furthermore, our analysis showed that the prevalence of a dialect in pre-training data more strongly influenced translation performance than its linguistic similarity to MSA, with Egyptian Arabic consistently outperforming other dialects. Our multi-stage Ara-TEaR method did not provide statistically significant gains and proved computationally expensive. The overall findings offer practical guidance and insights for prompt design in low-resource DA-MSA translation scenarios.

The resource-efficient fine-tuning approach established that task-specific fine-tuning can lead to state-of-the-art performance even with mid-sized models. The Gemma-2-9B model emerged as the best performing model, balancing high accuracy with computational efficiency, surpassing larger zero-shot models like GPT-4o. Key findings from our fine-tuning experiments also include:
\begin{itemize}
    \item Joint multi-dialect models outperformed dialect-specific ones due to shared linguistic features between dialects.
    \item Increasing training data volume and diversity significantly improved translation quality.
    \item Synthetic data augmentation without careful filtering negatively impacted performance, especially for larger models.
    \item The use of 4-bit quantization makes deploying these models in resource-constrained environments more practical by significantly reducing memory usage and improving inference speed with minimal quality loss.
\end{itemize}




\section*{Appendix}
\label{appendix:prompt-templates}


\section{Challenges and Limitations}
Despite the progress achieved, our work faced several limitations. Computational constraints restricted the scope of our experiments, limiting hyperparameter tuning and preventing fine-tuning of larger models. Resource restrictions also limited our data augmentation efforts. The focus on three dialects (Levantine, Gulf, Egyptian) constrained generalization, leaving the proposed translation techniques untested on varieties such as Maghrebi and Sudanese Arabic. Additionally, the overrepresentation of Egyptian Arabic in training data likely inflated performance for this dialect, prompting cautious interpretation of results.

Evaluation methodologies presented further challenges. Standard lexical metrics, such as SacreBLEU and CHrF++, are likely incapable of fully capturing meaning preservation in translations , especially with dialect-specific expressions. Comparing our fine-tuned models against prior works proved difficult due to the absence of public and standardized benchmark datasets and evaluation protocols, limiting the ability to make definitive performance claims.

These limitations highlight crucial areas for future work, including broader dialect coverage, improved evaluation metrics, and standardized benchmark development.

\section{Further Work}
Our future work could include investigating the performance of our models on a wider variety of dialects to identify any patterns and performance gaps. Refinements in data augmentation strategies are necessary to overcome the performance drop observed in our experiment. Future augmentation pipelines should integrate automatic filtering based on quality assessments to avoid training models on low-quality data. Given the current inability of language models to effectively back-translate from MSA to DA, alternative data augmentation techniques must be explored to enhance dataset diversity while maintaining high data quality.

Furthermore, a promising direction is developing specialized evaluation metrics for DA-MSA translation that align more closely with human assessment. These new automated metrics should account for dialect-specific linguistic features that are often overlooked by current automatic metrics, thereby advancing human-aligned automatic evaluation.

\section{Detailed Prompting Results}\label{sec:Detailed-Results}

\begin{table}[H]
\centering
\begin{tabular}{lcccc}
\multicolumn{5}{c}{Levantine Dialect - BLEU} \\
\hline
Model & 0-Shot & 0-Shot CoT & Few-Shot & Ara-TEaR \\
\hline
GPT-3.5 & 9.75 & 8.1 & 8.75 & 7.69 \\
GPT-4o & 11.98 & 12.03 & 12.94 & 11.32 \\
LLaMA 3.3-70B-Instruct & 6.64 & 5.89 & 6.42 & 7.52 \\
LLaMA 3.1-405B & 8.91 & 8.44 & 9.12 & 8.19 \\
Gemma-27B & 5.99 & 6.91 & 6.97 & 7.19 \\
DeepSeek-V3 & 10.52 & 10.86 & 10.57 & 10.56 \\
\hline
\end{tabular}
\caption{BELU scores comparing models performance with zero-shot, CoT, Ara-TEaR and Few-shot prompting strategies for Levantine Arabic to MSA translation.}
\label{tab:performance_comparison_lev}
\end{table}

\begin{table}[H]
\centering
\begin{tabular}{lcccc}
\multicolumn{5}{c}{Levantine Dialect - CHrF++} \\
\hline
Model & 0-Shot & 0-Shot CoT & Few-Shot & Ara-TEaR \\
\hline
GPT-3.5 & 35.85 & 33.94 & 34.12 & 33.67 \\
GPT-4o & 40.23 & 40.61 & 40.67 & 40.39 \\
LLaMA 3.3-70B-Instruct & 31.77 & 31.47 & 32.5 & 33.39 \\
LLaMA 3.1-405B & 35.19 & 34.33 & 34.91 & 34.31 \\
Gemma-27B & 32.28 & 32.79 & 32.35 & 33.28 \\
DeepSeek-V3 & 37.86 & 38.54 & 38.25 & 38.03 \\
\hline
\end{tabular}
\caption{CHrF++ scores comparing models performance with zero-shot, CoT, Ara-TEaR and Few-shot prompting strategies for Levantine Arabic to MSA translation.}
\label{tab:performance_comparison}
\end{table}

\begin{table}[htbp]
\centering
\begin{tabular}{lcccc}
\multicolumn{5}{c}{Gulf Dialect - BLEU} \\
\hline
Model & 0-Shot & 0-Shot CoT & Few-Shot & Ara-TEaR \\
\hline
GPT-3.5 & 9.47 & 8.56 & 7.95 & 6.91 \\
GPT-4o & 10.51 & 10.83 & 10.07 & 9.18 \\
LLaMA 3.3-70B-Instruct & 6.80 & 6.78 & 7.35 & 7.55 \\
LLaMA 3.1-405B & 7.37 & 8.43 & 9.14 & 7.38 \\
Gemma-27B & 7.26 & 7.74 & 8.03 & 8.62 \\
DeepSeek-V3 & 11.25 & 10.79 & 11.43 & 8.91 \\
\hline
\end{tabular}
\caption{BELU scores comparing models performance with zero-shot, CoT, Ara-TEaR and Few-shot prompting strategies for Gulf Arabic to MSA translation.}
\label{tab:performance_comparison_lev}
\end{table}

\begin{table}[htbp]
\centering
\begin{tabular}{lcccc}
\multicolumn{5}{c}{Gulf Dialect - CHrF++} \\
\hline
Model & 0-Shot & 0-Shot CoT & Few-Shot & TEaR \\
\hline
GPT-3.5 & 36.79 & 35.10 & 34.15 & 34.42 \\
GPT-4o & 38.78 & 39.17 & 38.53 & 37.38 \\
LLaMA 3.3-70B-Instruct & 33.23 & 32.40 & 33.84 & 33.56 \\
LLaMA 3.1-405B & 34.90 & 35.23 & 35.50 & 34.35 \\
Gemma-27B & 33.82 & 34.14 & 34.72 & 34.61 \\
DeepSeek-V3 & 38.54 & 37.84 & 38.87 & 36.73 \\
\hline
\end{tabular}
\caption{CHrF++ scores comparing models performance with zero-shot, CoT, Ara-TEaR and Few-shot prompting strategies for Gulf Arabic to MSA translation.}
\label{tab:performance_comparison_lev}
\end{table}

\begin{table}[htbp]
\centering
\begin{tabular}{lcccc}
\multicolumn{5}{c}{Egyptian Dialect - BLEU} \\
\hline
Model & 0-Shot & 0-Shot CoT & Few-Shot & Ara-TEaR \\
\hline
GPT-3.5 & 10.01 & 9.71 & 9.52 & 9.29 \\
GPT-4o & 12.19 & 12.27 & 13.32 & 11.76 \\
LLaMA 3.3-70B-Instruct & 10.59 & 9.34 & 9.73 & 9.87 \\
LLaMA 3.1-405B & 10.79 & 10.07 & 10.82 & 9.46 \\
Gemma-27B & 10.08 & 10.30 & 10.11 & 10.02 \\
DeepSeek-V3 & 12.47 & 12.57 & 12.49 & 10.52 \\
\hline
\end{tabular}
\caption{BELU scores comparing models performance with zero-shot, CoT, Ara-TEaR and Few-shot prompting strategies for Egyptian Arabic to MSA translation.}
\label{tab:performance_comparison_lev}
\end{table}

\begin{table}[htbp]
\centering
\begin{tabular}{lcccc}
\multicolumn{5}{c}{Egyptian Dialect - CHrF++} \\
\hline
Model & 0-Shot & 0-Shot CoT & Few-Shot & Ara-TEaR \\
\hline
GPT-3.5 & 38.69 & 38.38 & 37.27 & 37.36 \\
GPT-4o & 42.54 & 42.04 & 43.16 & 41.16 \\
LLaMA 3.3-70B-Instruct & 37.78 & 37.10 & 37.32 & 37.90 \\
LLaMA 3.1-405B & 38.51 & 38.12 & 38.84 & 37.38 \\
Gemma-27B & 37.29 & 38.09 & 37.87 & 37.44 \\
DeepSeek-V3 & 41.05 & 40.86 & 41.14 & 38.60 \\
\hline
\end{tabular}
\caption{CHrF++ scores comparing models performance with zero-shot, CoT, Ara-TEaR and Few-shot prompting strategies for Egyptian Arabic to MSA translation.}
\label{tab:performance_comparison_lev}
\end{table}

\section{Prompting Templates}\label{sec:prompting-templates}

\begin{tcolorbox}[
  colframe=black,
  colback=verylightgray,
  arc=5mm,
  boxrule=0.8pt,
  title=Zero-Shot Prompt Template
]
Translate the following text from \{dialect\} to Modern Standard Arabic: \{dialect\_text\}

Only output the final translation in Modern Standard Arabic; do not include any additional text.
\end{tcolorbox}

\begin{tcolorbox}[
  colframe=black,
  colback=verylightgray,
  arc=5mm,
  boxrule=0.8pt,
  title=Zero-Shot CoT Prompt Template
]
Translate the following text from \{dialect\} to Modern Standard Arabic: \{dialect\_text\}

Think step by step.

Only output the final translation in Modern Standard Arabic; do not include any additional text.

\end{tcolorbox}

\begin{figure}[H]
    \centering
    \includegraphics[width=1\linewidth]{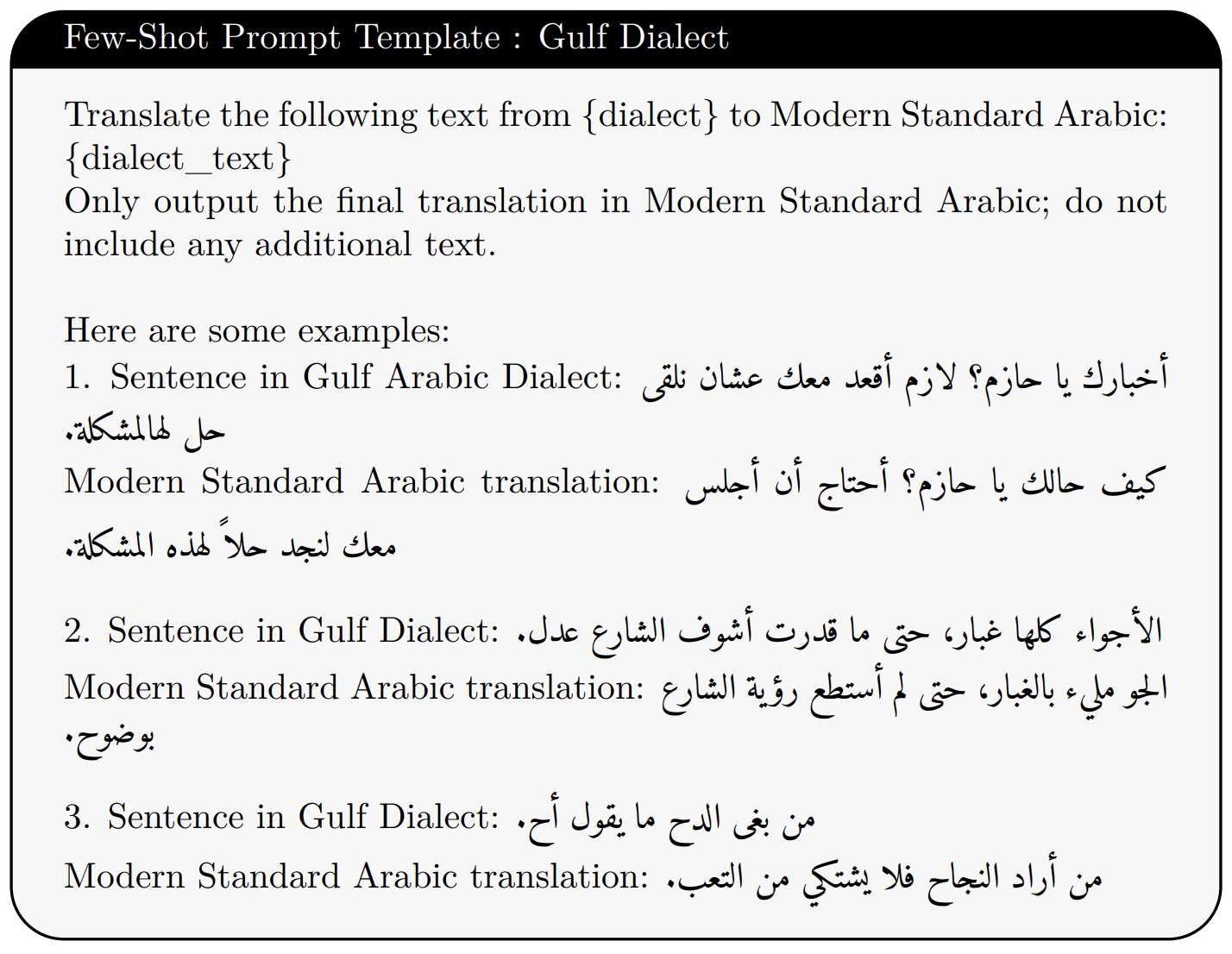}
    \label{fig:few-shot-gulf}
\end{figure}

\begin{figure}[H]
    \centering
    \includegraphics[width=1\linewidth]{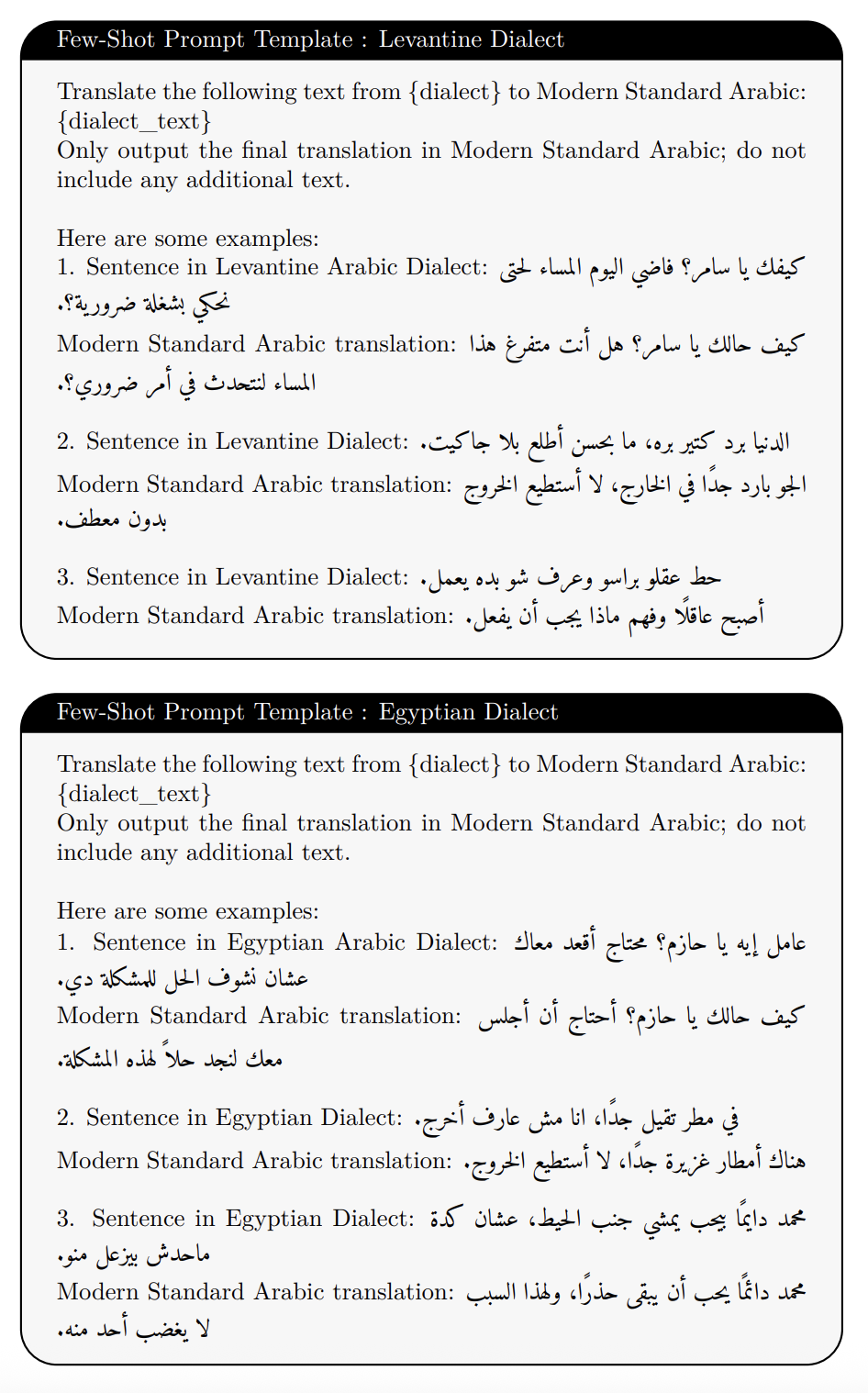}
    \label{fig:few-shot-lev-egy}
\end{figure}

\section*{Statements and Declarations}
\textbf{Conflict of interest}: The authors have no competing interests to declare.
\newline
\textbf{Funding}: No funding was received for conducting this study. 
\newline
\textbf{Data availability}: 
\begin{itemize}
    \item Our data and code are openly shared at Github: \newline  \url{https://github.com/er-abd/Advancing-DA2MSA-MT}
    \item Our fine-tuned models can be downloaded from Huggingface: \newline \url{https://huggingface.co/collections/er-abd/} (Advancing-DA2MSA-MT) 
\end{itemize}

\bibliographystyle{spbasic}
\bibliography{custom}

\end{document}